\documentclass[11pt,a4paper]{article}
\usepackage[hyperref]{emnlp2018}
\usepackage{times}
\usepackage{latexsym}

\usepackage{url}

\usepackage{amsmath}
\usepackage{dblfloatfix}
\usepackage{todonotes}
\usepackage{footnote}
\usepackage{flexisym}
\usepackage{multirow}
\usepackage{changepage}

\makesavenoteenv{align}
\makesavenoteenv{tabular}
\makesavenoteenv{table}
\makesavenoteenv{table*}
\usepackage{soul}
\usepackage{makecell}
\newenvironment{eqalign}{\par\nobreak\small\noindent\align}{\endalign}

\aclfinalcopy % Uncomment this line for the final submission

\title{Integrating Transformer and Paraphrase Rules \\ for Sentence Simplification}

\author{Sanqiang Zhao$^{\dagger}$
, Rui Meng$^{\dagger}$, Daqing He$^{\dagger}$, Saptono Andi$^{*}$, Parmanto Bambang$^{*}$ \\
  $^{\dagger}$ Department of Informatics and Networked Systems
, School of Computing and Information  \\
  $^{*}$ Department of Health Information Management, School of Health and Rehabilitation Sciences \\
  University of Pittsburgh \\
  Pittsburgh, PA, 15213 \\
  {\tt \{sanqiang.zhao, rui.meng, daqing, ans38, parmanto\}@pitt.edu} 
  \\}

\date{}

\begin{document}
\maketitle
\begin{abstract}
Sentence simplification aims to reduce the complexity of a sentence while retaining its original meaning. Current models for sentence simplification adopted ideas from machine translation studies and implicitly learned simplification mapping rules from normal-simple sentence pairs. In this paper, we explore a novel model based on a multi-layer and multi-head attention architecture and we propose two innovative approaches to integrate the Simple PPDB (A Paraphrase Database for Simplification), an external paraphrase knowledge base for simplification that covers a wide range of real-world simplification rules. The experiments show that the integration provides two major benefits: (1) the integrated model outperforms multiple state-of-the-art baseline models for sentence simplification in the literature (2) through analysis of the rule utilization, the model seeks to select more accurate simplification rules. The code and models used in the paper are available at \url{https://github.com/Sanqiang/text_simplification}.
  
%   two improved models, namely Deep Encouraged Sentence Simplification (DESS) model and Deep Memory Augmented Sentence Simplification (DMASS) model. Furthermore, we utilize the Simple PPDB, an external paraphrase knowledge base for simplification that covers a wide range of real-world simplification rules, to make models learn how to simplify directly. Our models outperform multiple state-of-the-art baselines for sentence simplification in the literature. Through analyzing the rule utilization, our model, especially DMASS, achieves the best in balancing the tendency of applying as many simplification rules as possible and the accuracy of applying the appropriate rules. The code and models used in the paper are available at \url{*}.
\end{abstract}

\section{Introduction}

Sentence simplification aims to reduce the complexity of a sentence while retaining its original meaning. It can benefit individuals with low-literacy skills ~\cite{watanabe2009facilita} including children, non-native speakers and individuals with language impairments such as dyslexia~\cite{rello2013dyswebxia}, aphasic~\cite{carroll1999simplifying}.

Most of the previous studies tackled this task in a way similar to machine translation~\cite{xu2015show, zhang2017sentence}, in which models are trained on a large number of pairs of sentences, each consisting of a normal sentence and a simplified sentence. Statistical and neural network modeling are two major methods used for this task. The statistical models have the benefit of easily integrating with human-curated rules and features, thus they generally perform well even they are trained with a limited number of data. In contrast, neural network models could learn the simplifying rules automatically without the need for feature engineering, but at the cost of requiring a huge amount of training data. Even though models based on neural networks have outperformed the statistical methods in multiple Natural Language Processing (NLP) tasks, their performance in sentence simplification is still inferior to that of statistical models~\cite{xu2015show, zhang2017sentence}. We speculate that current training datasets may not be large and broad enough to cover common simplification situations. However, human-created resources do exist which can provide abundant knowledge for simplification. This motivates us to investigate if it is possible to train neural network models with these types of resources.

Another limitation to using existing neural network models for sentence simplification is that they are only able to capture frequent transformations; they have difficulty in learning rules that are not frequently observed despite their significance. This may be due to nature of neural networks~\cite{feng2017memory}: during training, a neural network tunes its parameters to learn how to simplify different aspects of the sentence, which means that all the simplification rules are actually contained in the shared parameters. Therefore, if one simplification rule appears more frequently than others, the model will be trained to be more focused on it than the infrequent ones. Meanwhile, models tend to treat infrequent rules as noise if they are merely trained using sentence pairs. If we can leverage an additional memory component to maintain simplification rules individually, it would prevent the model from forgetting low-frequency rules as well as help it to distinguish real rules from noise. Therefore, we propose the Deep Memory Augmented Sentence Simplification (DMASS) model. 
For comparison purpose, we also introduce another approach, Deep Critic Sentence Simplification (DCSS) model, to encourage applying the less frequently occurring rules by revising the loss function.
It this way, simplification rules are encouraged to maintained internally in the shared parameters while avoiding the consumption of an unwieldy amount of additional memory.

In this study, we propose two improvements to the neural network models for sentence simplification. For the first improvement, we propose to use a multi-layer, multi-head attention architecture~\cite{vaswani2017attention}. Compared to RNN/LSTM (Recurrent Neural Network / Long Short-term Memory), the multi-layer, multi-head attention model would be able to selectively choose the correct words in the normal sentence and simplify them more accurately. 

Secondly, we propose two new approaches to integrate neural networks with human-curated simplification rules. Note that previous studies rarely tried to incorporate explicit human language knowledge into the encoder-decoder model.
Our first approach, DMASS, maintains additional memory to recognize the context and output of each simplification rules. Our second approach, DCSS, follows a more traditional approach to encode the context and output of each simplification rules into the shared parameters.

Our empirical study demonstrates that our model outperforms all the previous sentence simplification models. 
They achieve both a good coverage of rules to be applied (recall) and a high accuracy gained by applying the correct rules (precision).

\section{Related Work}

\paragraph{Sentence Simplification}

For statistical modeling, \citet{zhu2010monolingual} proposed a tree-based sentence simplification model drawing inspiration from statistical machine translation. \citet{woodsend2011learning} employed quasi-synchronous grammar and integer programming to score the simplification rules. \citet{wubben2012sentence} proposed a two-stage model PBMT-R, where a standard phrase-based machine translation (PBMT) model was trained on normal-simple aligned sentence pairs, and several best generations from PBMT were re-ranked based how dissimilar they were to a normal sentence. Hybrid, a model proposed by \citet{narayan2014hybrid} was also a two-stage model combining a deep semantic analysis and machine translation framework. 
SBMT-SARI \cite{xu2016optimizing} achieved state-of-the-art performance by employing an external knowledge base to promote simplification.
In terms of neural network models, \citet{zhang2017sentence} argued that the RNN/LSTM model generated sentences but it does not have the capability to simplify them. They proposed DRESS and DRESS-LS that employ reinforcement learning to reward simpler outputs. As they indicated, the performance is still inferior due to the lack of external knowledge.
Our proposed model is designed to address the deficiency of current neural network models which are not able to integrate an external knowledge base.

\paragraph{Augmented Dynamic Memory}

Despite positive results obtained so far, a particular problem with the neural network approach is that it has a tendency towards favoring to frequent observations but overlooking special cases that are not frequently observed. This weakness with regard to infrequent cases has been noticed by a number of researchers who propose an augmented dynamic memory for multiple applications, such as language models~\cite{daniluk2017frustratingly, grave2016improving}, question answering~\cite{miller2016key}, and machine translation~\cite{feng2017memory, tu2017learning}. We find that current sentence simplification models suffer from a similar neglect of infrequent simplification rules, which inspires us to explore augmented dynamic memory.

\section{Our Sentence Simplification Models}

\subsection{Multi-Layer, Multi-Head Attention}

\begin{figure}[h]
\begin{center}
   \includegraphics[width=1.0\linewidth]{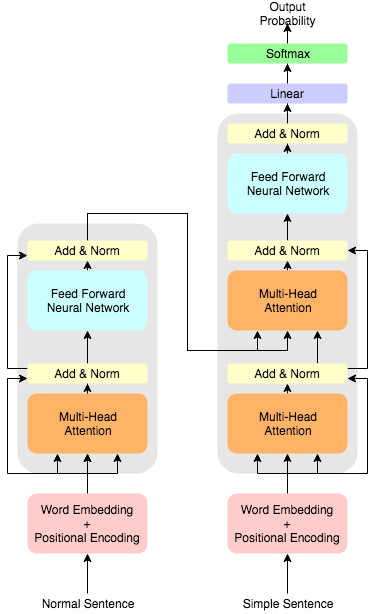}
\end{center}
\vspace{-0.5em}
   \caption{Diagram of the Transformer architecture}
\label{fig:transformer}
\vspace{-0.5em}
\end{figure}

Our basic neural network-based sentence simplification model utilizes a multi-layer and multi-head attention architecture~\cite{vaswani2017attention}. 
As shown in Figure~\ref{fig:transformer}, our model based on the Transformer architecture works as follows: given a pair consisting a normal sentence $I$ and a simple sentence $O$, the model learns the mapping from $I$ to $O$. 

The encoder part of the model (see the left part of Figure~\ref{fig:transformer}) encodes the normal sentence with a stack of $L$ identical layers. Each layer has two sublayers: one layer is for multi-head self-attention and the other one is a fully connected feed-forward neural network for transformation.
The multi-head self-attention layer encodes the output from the previous layer into hidden state $e_{(s,l)}$ (step $s$ and layer $l$) as shown in Equation \ref{eq:encoder1}, where $\alpha_{(s',l)}^{enc}$ indicates the attention distribution over the step $s'$ and layer $l$.
Each hidden state summarizes the hidden states in the previous layer 
through the multi-head attention function $a()$~\cite{vaswani2017attention} where H refers to the number of heads.

The right part of Figure~\ref{fig:transformer} denotes the decoder for generating the simplified sentence. The decoder also consists of a stack of $L$ identical layers. In addition to the same two sub-layers as those in the encoder part, the decoder also inserts another multi-head attention layer aiming to attend on the encoder outputs. The bottom multi-head self-attention plays the same role as the one in the encoder, where the hidden state $d_{(s,l)}$ is computed in the Equation~\ref{eq:decoder1}. The upper multi-head attention layer is used to seek relevant information from encoder outputs. Through the same mechanism, context vector $c_{(s,l)}$ (step $s$ and layer $l$) is computed in the Equation~\ref{eq:decoder2}.

\begin{eqalign}
\fontsize{9}{11}\selectfont
\label{eq:encoder1}
  e_{(s,l)} = & \sum_{s'} \alpha_{(s',l)}^{enc} e_{(s',l-1)},&\alpha_{(s',l)}^{enc} = & a(e_{(s,l)}, e_{(s', l-1)}, H)\footnote{The lowest hidden state $e_{(:, 0)}$ is the word embedding.} \\ 
  d_{(s,l)} = & \sum_{s_\textprime} \alpha_{(s_\textprime,l)}^{dec} d_{(s_\textprime,l-1)},
\label{eq:decoder1}
 &  \alpha_{(s_\textprime,l)}^{dec} = & a(d_{(s,l)}, c_{(s_\textprime, l-1)}, H)\footnote{The lowest context vector $c_{(:, 0)}$ is the word embedding.} \\
\label{eq:decoder2}
  c_{(s,l)} = & \sum_{s_\textprime} \alpha_{(s_\textprime,l)}^{dec2} e_{(s_\textprime,L)}, 
  & \alpha_{(s_\textprime,l)}^{dec2} = & a(d_{(s,l)}, e_{(s_\textprime,L)}, H) 
\end{eqalign}
\vspace{-1.0em}

The model is trained to minimize the negative log-likelihood of the simple sentence, $L_{seq} = -logP(O|I, \theta)$ where $\theta$ represents all the parameters in the current model.

\subsection{Integrating with Simple PPDB}
A previous study~\cite{xu2016optimizing} has demonstrated the benefits of using an external knowledge base in conjunction with a statistical simplification model. However, as far as we know, no efforts have been made to integrate neural network models with the knowledge base, and our study is the first to meet this goal.

\begin{table}[h!]
% \fontsize{9}{11}\selectfont
\small
\begin{center}
\begin{tabular}{|l|l|l|l|}
\hline  Weight & Type & Rule \\ \hline
 0.99623& [VP]  &recipient $\rightarrow$ have receive \\
 0.75530& [NN]  &recipient $\rightarrow$ winner \\
 0.58694& [NN]  &recipient $\rightarrow$ receiver \\
 0.46935& [NN]  &recipient $\rightarrow$ host \\
\hline
\end{tabular}
\end{center}
\caption{Examples from the Simple PPDB}
\label{tab:simpleppdb}
\end{table}

Simple PPDB~\cite{pavlick2016simple} refers to a paraphrase knowledge base for simplification. It is a refined version of another knowledge, PPDB~\cite{ganitkevitch2013ppdb}, which was originally designed to support paraphrase. Simple PPDB contains 4.5 million paraphrase rules, each of which provides the mapping from a normal phrase to a simplified phrase, the syntactic type of the normal phrase, and the simplification weight. Table~\ref{tab:simpleppdb} shows four examples, where ``recipient'' can be simplified to ``winner'' with a weight 0.75530 if ``recipient'' is a singular noun (NN). 
% or simplified to ``have received'' if ``recipient'' is a verb phrase (VP) with weight 0.99623.

% We pre-process the training dataset to identify the correct rules to be applied for each normal-simple sentence pairs in training dataset.

\subsubsection{Deep Critic Sentence Simplification Model (DCSS)}

The Simple PPDB offers guidance about whether a word needs to be simplified and how it should be simplified. The Deep Critic Sentence Simplification (DCSS) model is designed to apply rules identified by the Simple PPDB by introducing a new loss function. Different from the standard loss function that minimizes the distance away from the ground truth, the new loss function aims to maximize the likelihood of applying simplification rules. It also reweights the probability of generating each word by its simplification weight in order to relieve the problem of overlooking infrequent simplification rules.

For example, given a normal sentence in the training set, ``the recipient of the kate greenaway medal'', the simplified sentence is ``the winner of the kate greenaway medal.'', where ``recipient'' is simplified to ``winner'', which is identified by Simple PPDB. The major goal of the loss functions is to support the model in generating the simplified word ``winner'' while deterring the model from generating the word ``recipient''. Specifically, for an applicable simplification rule, our new loss function maximizes the probability of generating the simplified form (word ``winner'') and meanwhile minimizes the probability of generating the original form (word ``recipient''). As in Equation \ref{eq:critic_loss}, where $w_{rule}$ indicates the weight of the simplification rule provided by the Simple PPDB, once the model generates ``recipient'', the model is criticized to generate word ``winner''; when model predicts correctly with ``winner'', the model is trained to minimize the probability of ``recipient''.
In this way, the model avoids selecting normal words and instead becomes inclined to choose the simplified words.

\begin{eqalign}
\fontsize{9}{11}\selectfont
\small
\label{eq:critic_loss}
\begin{split}
L_{critic} =
\begin{cases}
-w_{rule} log P(winner|I,\theta)& \\ \hspace{20 mm}  \text{if model generates recipient}\\
w_{rule} log P(recipient|I,\theta)& \\ \hspace{20 mm}   \text{if model generates winner}
\end{cases}
\end{split}
\end{eqalign}

The $L_{critic}$ merely focuses on the words identified by the Simple PPDB and $L_{seq}$ focuses on the entire vocabulary. So, the model is trained in an end-to-end fashion by minimizing $L_{seq}$ and $L_{critic}$ alternately.

\subsubsection{Deep Memory Augmented Sentence Simplification Model (DMASS)}
%\todo{I rewrote this paragraph a lot. you need to check. }
%\todo{I just rewrote the reason why we change from DESS to DMASS. }
% DESS can potentially introduce grammatical mistakes by applying inappropriate rules. To overcome this problem, we think that a good model should remember the context where the rule should be applied.
DCSS, similar to the majority of neural network models, uses a piece of shared memory, i.e. the parameters, as the media to store the learned rules from the data.
%\todo{the rules it learned? the rules that applied in the training?}
%\todo{Sanqiang: same as above, rename it simplification rules that I think it is more clear.}
As a result, it still focuses much more on rules that are frequently observed and ignores the rules observed infrequently. 
However, infrequent rules are still important, particularly when the training data is limited. 

In order to make full use of the rules in the knowledge base, we introduce the Deep Memory Augmented Sentence Simplification (DMASS) model. DMASS has an augmented dynamic memory to record multiple key-value pairs for each rule in the Simple PPDB. The key vector stores a context vector that is computed based on the weighted average of encoder hidden states and the current decoder hidden states. The value vector stores the output vector.

\begin{figure}[h]
\begin{center}
   \includegraphics[width=1.0\linewidth]{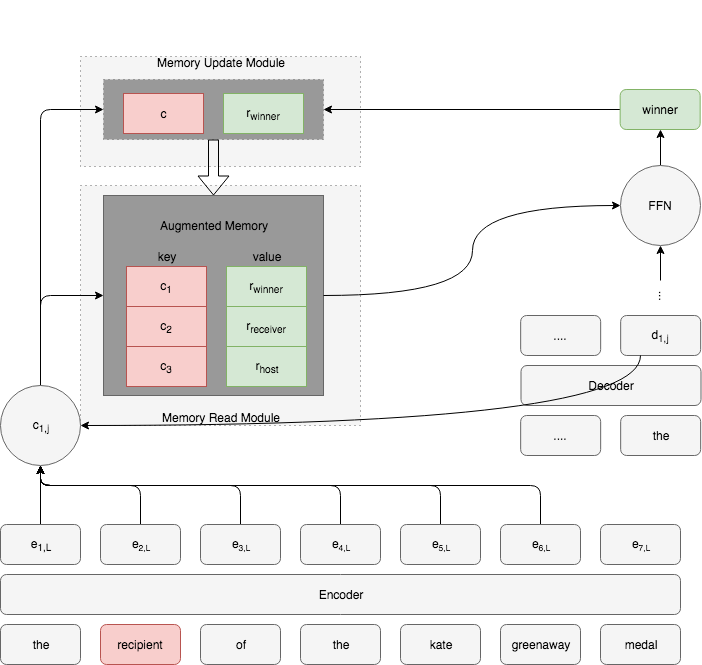}
\end{center}

\vspace{-0.5em}
\caption{Diagram of DMASS Model}
\label{fig:augmented_memory}
\vspace{-0.5em}
\end{figure}

Our DMASS model is illustrated in Figure~\ref{fig:augmented_memory}. Given the same example normal sentence `` the recipient of kate greenaway medal'', Simple PPDB determines that the word ``recipient'' should be simplified to ``winner''. 
The encoder represents the normal sentence as a list of hidden states, [$e_{(1,L)}$, $e_{(2,L)}$, ...] where L indicates the final layer of encoder hidden states. When predicting the next word in the simplified sentence, the decoder of layer $j$ represents the previous words as hidden states [$d_{(1,j)}$, ... ]. $c_{(1,j)}$ refers to the current context vector following attention layer, which is the weighted average of [$e_{(1,L)}$, $e_{(2,L)}$, ...] based on $d_{(1,j)}$. 
A feed-forward fully connected neural network (FFN) combines the output of the decoder and the output from memory read module into the final output $r_{winner}$. In addition to the word prediction, $c_{(1,j)}$ and $r_{winner}$ will be sent to memory update module.

In the remainder of this section, we will introduce the two modules of DMASS mentioned above:
%augmented memory 
Memory Read Module and Memory Update Module.

% \todo{should the caption be changed to ``Diagram of DMASS Model''? }

\vspace{-0.5em}

\paragraph{Memory Read Module}

The memory read module incorporates rules into prediction. As shown in Figure~\ref{fig:augmented_memory}, current augmented memory contains three candidate rules for the word ``recipient'', which indicates that it can be simplified into ``winner'', ``receiver'' or ``host'', respectively. The current context vector $c_{(1,j)}$ is treated as a query to search for suitable rules by using Equation \ref{eq:memory_read1}, where $\alpha_{i}^{r}$ denotes the weight for $i^{th}$ rule, which is computed through the dot product between current context vector $c_{(1,j)}$ and $c_i$. Then using Equation \ref{eq:memory_read2}, $\alpha_{i}^{r}$ weights each output vector to generate memory read output.

\begin{eqalign}
\fontsize{9}{11}\selectfont
\label{eq:memory_read1}
  \alpha_{i}^{r} = &\frac{e_{i}}{\sum_j e_{j}} &e_{i} &= exp(c_{(1,j)} \cdot c_{ i}) \\
\label{eq:memory_read2}
  r_{o} = &\sum \alpha_{i}^{r} r_{r} &r_{r} &\in [r_{winner}, r_{receiver}, r_{host}] 
\end{eqalign}

\paragraph{Memory Update Module}

The task of the memory update module is to update the key and value vectors in the augmented memory. Once the model predicts the output vector $r_{winner}$, both $r_{winner}$ and the current context vector $c_{1,j}$ are sent to the memory update module. If the augmented memory does not contain the key-value pair for the rule, $c_{1,j}$ and $r_{winner}$ are appended to the memory. If the augmented memory contains the key-value pair, the key vector is updated as the mean of current key vector and $c_{1,j}$. Similarly, the value vector is also updated as the mean of current value vector and $r_{winner}$.
% the updated key and value vector is the average of key-value vector and the coming $c$-$r_{winner}$ vector respectively.

% The training of DMASS contains three stages. In the first stage (the first 50,000 global steps), the model is trained without augmented memory module for warm-up. Then in the second stage (next 50,000 global steps), 
%\todo{not know what this sentence tries to say. }
%\todo{Sanqiang: in the first stage, model is only pretrained (we want the trained vector going into the augmented memory); the second stage we turn on the module update module that we populate the augmented memory (just populate the augmented memory); the third stage we turn on both memory update module and memory read module (we use augmented memory as well as update augmented memory)}
% the model turns on the memory update module that enables the population of the memory. Lastly, the model is trained by turning on both memory read and update module.

\section{Experiments}

% In order to examine the performance of our models, we design a set of experiments. To make the results comparable, our experimental setup is same to previous studies~\cite{xu2016optimizing, zhang2017sentence}.

\paragraph{Dataset}

We utilize the dataset \textit{WikiLarge}~\cite{zhang2017sentence} for training. It is the largest Wikipedia corpus, constructed by merging previously created simplification corpora. Specifically, the training dataset contains 296,402 normal-simple sentence pairs gathered from~\cite{zhu2010monolingual, woodsend2011learning, kauchak2013improving}.
For validation and testing, we use the dataset \textit{Turk} created by~\cite{xu2016optimizing}. In this dataset, eight simplified reference sentences for each normal sentence are used as the ground-truth, all of which are generated by Amazon Mechanical Turk workers. The \textit{Turk} dataset contains 2,000 data samples for validation and 356 samples for testing. We consider the \textit{Turk} to be the most reliable data set because (1) it is human-generated and (2) it contains multiple simplification references for each normal sentence due to the existence of multiple equally good simplifications of each sentence.
We also include the second test set \textit{Newsela}, a corpus introduced by ~\cite{xu2015problems} who argue that only using normal-simple sentence pairs from Wikipedia is suboptimal due to the automatic sentence alignment which unavoidably introduces errors, and the uniform writing style which leads to systems that generalize poorly. 
The test set contains 1,419 normal-simple sentence pairs\footnote{Because the earlier publications don't provide pre-process details, we use our own script to pre-process the articles into sentence pairs.}. To demonstrate that our models are able to perform well on a different style of corpus, we report the results of \textit{Newsela} test set by using the models trained/tuned on \textit{Turk} dataset.
Following~\citet{zhang2017sentence}'s way, we tag and anonymize name entities with a special token in the format of \textbf{NE@N}, where NE includes $\left\{PER,LOC,ORG \right\}$ and N indicates the $N^{th}$ distinct NE type of entity. We also replace those tokens occurring three times or less in the training set with a mark ``UNK'' as mentioned in~\cite{zhang2017sentence}.

\paragraph{Evaluation Metrics}

We report the results of the experiment with two metrics that are widely used in the literature: SARI~\cite{xu2016optimizing} and FKGL~\cite{kincaid1975derivation}.
% BLEU measures the degree of a simplified sentence being different from the ground truth references. The higher value of BLEU is, the better match between the simplified sentence and the ground truth references is. 
FKGL computes the sentence length and word length as a way to measure the simplicity of a sentence. The lower value of FKGL indicates simpler sentence.
% In contrast,
FKGL measures the simplicity of a sentence without considering the ground truth simplification references and it correlates little with human judgment~\cite{xu2016optimizing}, so we also use another metric, SARI.
SARI, which stands for ``System output Against References and against the normal sentence'', computes the arithmetic mean of N-grams (N includes 1,2,3 and 4) F1-score of three rewrite operations: addition, deletion, and keeping.
% \todo{you call three operations addition, deletion and copying, but later int the result section, you just call them add, delete, and keep. maybe you want to be consistent on calling them. }
Specifically, it rewards addition operations where a word in the generated simplified sentence does not appear in the normal sentence but is mentioned in the reference sentences. It also rewards words kept or deleted in both the simplified sentence and the reference sentences. In our experiment, we also present the F1-score of three rewrite operations: addition, deletion, and keeping.
\citet{xu2016optimizing} demonstrated that SARI correlates most closely to human judgments in sentence simplification tasks. Thus, we treated SARI as the most important measurement in our study.
% We reuse the script from \cite{xu2016optimizing} to compute the corpus-level multi-reference SARI and FKGL.

Because SARI rewards deleting and adding separately, we also include another metric to measure the correctness of lexical transformation, namely word simplification, verified by Simple PPDB. By comparing the normal sentence and ground truth simplified references, we collect rules that are correct to be used for simplifying each normal sentence. Then we calculate the precision, recall, and F1 score for using the correct rules.  As a result, the recall expresses the coverage of rules to be applied, and the precision implies the accuracy gained by applying the correct rules.

\paragraph{Training Details}

%\todo{you need to at least give an overview of what you have done with training. for example, which part of the dataset was used for training}
%\todo{I rephrase the Dataset paragraph to identify training/evaluation/test}
We initialized the encoder and decoder word embedding lookup matrices with 300-dimensional Glove vectors\cite{pennington2014glove}. The word embedding dimensionality and the number of hidden units are set to 300. During the training, we regularize all layers with a dropout rate of 0.2~\cite{srivastava2014dropout}. For multi-layer and multi-head architecture, 4 encoder and decoder layers (set L as 4) and 5 multi-attention heads (set H as 5) are used. We will discuss the trade-off between different layers and different heads in Sections~\ref{sec:result-multilayermultihead}.
For DMASS, we use the context vector based on the first layer of the decoder (set j as 1).
For optimization, we use Adagrad~\cite{duchi2011adaptive} with the learning rate set to 0.1. The gradient is truncated by 4~\cite{pascanu2013difficulty}. % All these training activities are performed on a single NVIDIA GPU card.

\subsection{Impacts of Multi-Layer, Multi-Head Attention Architecture}
\label{sec:result-multilayermultihead}

The reason to employ the Transformer architecture in the sentence simplification task is that we believe that its multi-layer, multi-head attention provides a better capability of modeling both the overall context and the important cues for sentence simplification. In this section, we examine the applicability of multi-layer, multi-head attention architecture to the sentence simplification task. We compare our results against the RNN/LSTM-based sentence simplification models. Note that the results of our models presented here have not been integrated with the Simple PPDB.

\begin{table*}[!h]
\fontsize{8}{8}\selectfont
\centering
\renewcommand{\arraystretch}{1.3}
\begin{tabular}{|l|l|ll|l|lll|}
\hline
\multicolumn{1}{|c|}{\multirow{2}{*}{Model}} & \multicolumn{1}{c|}{\multirow{2}{*}{FKGL}} & \multicolumn{2}{c|}{Factors in FKGL} & \multicolumn{1}{c|}{\multirow{2}{*}{SARI}} & \multicolumn{3}{c|}{F1 for operations in FKGL} \\ \cline{3-4} \cline{6-8} 
\multicolumn{1}{|c|}{} & \multicolumn{1}{c|}{} & \multicolumn{1}{c|}{WLen} & \multicolumn{1}{c|}{SLen} & \multicolumn{1}{c|}{} & \multicolumn{1}{c|}{Add} & \multicolumn{1}{c|}{Delete} & \multicolumn{1}{c|}{Keep} \\ \hline
RNN/LSTM & 8.67 & 1.34 & 21.68 & 35.66 & 3.00 & 28.95 & \underline{75.03} \\
Transformer (L1H5) & 8.59 & 1.34 & 21.39 & 35.88 & 2.69 & 30.46 & 74.50 \\
Transformer (L2H5) & 8.11 & 1.33 & 20.52 & 36.88 & 3.48 & 33.26 & 73.91 \\
Transformer (L3H5) & 7.71 & 1.32 & 19.77 & 38.02 & 4.14 & 37.41 & 72.51 \\
Transformer (L4H1) & 7.49 & 1.31 & 19.41 & 37.88 & 4.05 & 37.35 & 72.23 \\
Transformer (L4H2) & 7.40 & 1.31 & 19.19 & 38.35 & 4.58 & 39.77 & 70.70 \\
Transformer (L4H5) & \underline{7.22} & \underline{1.30} & \underline{19.00} & \underline{38.84} & \underline{4.78} & \underline{41.19} & 70.53 \\ \hline
\end{tabular}
\caption{Comparison of transformers with different layers and heads of attention on Turk dataset}
\label{tab:trans_layer}
\end{table*}

Table~\ref{tab:trans_layer} shows the experiment results where LxHy indicates a run with Transformer using $x$ layers and $y$ heads. When compared with results of RNN/LSTM, our Transformer-based model performed better in terms of SARI and FKGL values.
In addition, with the increased number of layers or heads, the values of SARI and FKGL improve accordingly.
In the remainder of this section, we analyze the insights of these results in detail.

In our tasks, FKGL measures the sentence length and the word length as two factors for evaluating a simplified sentence. Therefore, we include \textit{Wlen(Word Length)} and \textit{Slen(Sentence Length)} into our analysis. As shown in Table~\ref{tab:trans_layer}, models with higher numbers of layers and/or heads do generally reduce the average word length and the average sentence length, which indicates that the higher number of layers and/or heads in the model leads to simpler outcomes.

It has been found that SARI correlates most closely to human judgment~\cite{xu2016optimizing}. To further analyze the effects of SARI, we study the impacts of three rewrite operations in SARI: add, delete, and keep. As shown in Table~\ref{tab:trans_layer}, we find that the improvement mostly results from correctly adding simplified words and deleting normal words, but not from keeping words. By analyzing the outputs, the increased number of layers or heads results in better capability to simplify the words. Specifically, models with the greater number of layers or heads tend to remove the normal words and add simplified words. However, they may introduce inaccurate simplified words, thereby driving down the F1 score for keeping words. We believe the Simple PPDB, which offers guidance about whether words need to be simplified and how they should be simplified, provides an ideal method to alleviate this issue.

\subsection{Impacts of Integrating the Simple PPDB}
\label{sec:result-simpleppdb}

\begin{table*}[!h]
\fontsize{8}{8}\selectfont
\centering
\renewcommand{\arraystretch}{1.3}
% \begin{adjustwidth}{-0.7cm}{}
\begin{tabular}{|l|l|ll|l|lll|lll|}
\hline
\multirow{2}{*}{Model} & \multicolumn{1}{c|}{\multirow{2}{*}{FKGL}} & \multicolumn{2}{c|}{Factors in FKGL} & \multicolumn{1}{c|}{\multirow{2}{*}{SARI}} & \multicolumn{3}{c|}{F1 for operations of SARI} & \multicolumn{3}{c|}{Rule Utilization} \\ \cline{3-4} \cline{6-11} 
 & \multicolumn{1}{c|}{} & \multicolumn{1}{l|}{WLen} & SLen & \multicolumn{1}{c|}{} & \multicolumn{1}{l|}{Add} & \multicolumn{1}{l|}{Delete} & Keep & \multicolumn{1}{l|}{Prec} & \multicolumn{1}{l|}{Recall} & F1 \\ \hline
PBMT-R & 8.35 & 1.30 & 22.08 & \multicolumn{1}{c|}{38.56} & 5.73 & 36.93 & 73.02 & 14.60 & 22.29 & 15.01 \\
Hybrid & \underline{4.71} & 1.28 & \underline{13.38} & \multicolumn{1}{c|}{31.40} & 5.49 & 45.48 & 46.86 & 10.62 & 7.61 & 7.62 \\
SBMT-SARI & 7.49 & \underline{1.18} & 23.50 & \multicolumn{1}{c|}{39.96} & \underline{5.97} & 41.43 & 72.51 & 13.30 & 28.96 & 15.77 \\
RNN/LSTM & 8.67 & 1.34 & 21.68 & \multicolumn{1}{c|}{35.66} & 3.00 & 28.95 & \underline{75.03} & 13.67 & 14.83 & 11.65 \\
DRESS & \underline{6.80} & 1.34 & \underline{16.55} & \multicolumn{1}{c|}{37.08} & 2.94 & 43.14 & 65.16 & 13.06 & 12.50 & 10.77 \\
DRESS-LS & \underline{6.92} & 1.35 & \underline{16.76} & \multicolumn{1}{c|}{37.27} & 2.82 & 42.21 & 66.78 & 12.40 & 11.36 & 9.83 \\ \hline
DMASS & 7.41 & 1.29 & 20.00 & 39.81 & 5.04 & 41.94 & 72.46 & 17.97 & 25.54 & 18.12 \\
DCSS & 7.34 & 1.31 & 19.30 & 39.26 & 5.29 & 41.24 & 71.26 & 13.14 & 21.30 & 13.87 \\
DMASS+DCSS & 7.18 & 1.27 & 20.10 & 40.42 & 5.48 & \underline{45.55} & 70.22 & 16.25 & 30.42 & 18.98 \\ \hline
DMASS$_{beam=4}$ & 8.20 & 1.30 & 21.66 & 39.16 & 4.90 & 38.41 & 74.18 & 18.53 & 25.46 & 18.40 \\
DCSS$_{beam=4}$ & 7.97 & 1.32 & 20.56 & 39.11 & 5.10 & 38.87 & 73.36 & 14.36 & 20.96 & 14.48 \\
DMASS+DCSS$_{beam=4}$ & 7.93 & 1.28 & 21.49 & 40.34 & 5.73 & 42.55 & 72.74 & 18.55 & 31.56 & 20.81 \\ \hline
DMASS$_{beam=8}$ & 8.23 & 1.30 & 21.68 & 39.15 & 4.95 & 37.80 & 74.69 & 18.44 & 25.34 & 18.32 \\
DCSS$_{beam=8}$ & 7.97 & 1.32 & 20.56 & 39.11 & 5.10 & 38.87 & 73.36 & 14.37 & 20.96 & 14.80 \\
DMASS+DCSS$_{beam=8}$ & 8.04 & 1.29 & 21.64 & \underline{40.45} & 5.72 & 42.23 & 73.41 & \underline{19.46} & \underline{31.99} & \underline{21.51} \\ \hline
\end{tabular}
% \end{adjustwidth}
\caption{Performance of baselines and proposed models on the Turk dataset.}
\label{tab:perf_all}
\end{table*}

\begin{table*}[!h]
\fontsize{9}{9}\selectfont
\centering
\renewcommand{\arraystretch}{1.3}
% \begin{adjustwidth}{-2.5cm}{}
\begin{tabular}{|l|l|ll|l|lll|lll|}
\hline
\multirow{2}{*}{Model} & \multicolumn{1}{c|}{\multirow{2}{*}{FKGL}} & \multicolumn{2}{c|}{Factors in FKGL} & \multicolumn{1}{c|}{\multirow{2}{*}{SARI}} & \multicolumn{3}{c|}{F1 for operations of SARI} & \multicolumn{3}{c|}{Rule Utilization} \\ \cline{3-4} \cline{6-11} 
 & \multicolumn{1}{c|}{} & \multicolumn{1}{l|}{WLen} & SLen & \multicolumn{1}{c|}{} & \multicolumn{1}{l|}{Add} & \multicolumn{1}{l|}{Delete} & Keep & \multicolumn{1}{l|}{Prec} & \multicolumn{1}{l|}{Recall} & F1 \\ \hline
RNN/LSTM & 6.09 & 1.22 & 18.67 & \multicolumn{1}{c|}{21.09} & 11.10 & 38.78 & 13.39 & 12.62 & 22.63 & 14.68 \\
DRESS & \underline{4.96} & 1.23 & \underline{15.27} & \multicolumn{1}{c|}{25.70} & 10.65 & 52.59 & 13.86 & 12.56 & 17.88 & 13.28 \\
DRESS-LS & \underline{5.07} & 1.24 & \underline{15.47} & \multicolumn{1}{c|}{24.91} & 11.21 & 49.74 & 13.76 & 12.61 & 17.50 & 13.42 \\ \hline
DMASS & 5.38 & 1.20 & 17.47 & 25.41 & 11.88 & 50.39 & 13.97 & 16.32 & 34.79 & 20.00 \\
DCSS & 5.64 & 1.22 & 17.58 & 24.31 & 13.52 & 45.60 & 13.81 & 15.20 & 30.38 & 18.39 \\
DMASS+DCSS & 5.17 & \underline{1.18} & 17.60 & \underline{27.28} & 11.56 & \underline{56.10} & \underline{14.19} & 15.98 & 40.64 & 20.98 \\ \hline
DMASS$_{beam=4}$ & 5.64 & 1.21 & 17.79 & 24.09 & 13.96 & 44.47 & 13.85 & 17.40 & 35.97 & 21.37 \\
DCSS$_{beam=4}$ & 5.80 & 1.22 & 17.85 & 23.28 & 15.28 & 40.76 & 13.81 & 16.77 & 31.81 & 20.06 \\
DMASS+DCSS$_{beam=4}$ & 5.42 & 1.19 & 17.81 & 26.39 & 13.92 & 51.13 & 14.13 & 18.71 & 43.36 & 24.23 \\ \hline
DMASS$_{beam=8}$ & 5.68 & 1.21 & 17.83 & 23.95 & 14.25 & 43.74 & 13.86 & 17.69 & 36.37 & 21.74 \\
DCSS$_{beam=8}$ & 5.77 & 1.22 & 17.76 & 23.18 & \underline{15.65} & 40.08 & 13.82 & 17.18 & 32.18 & 20.50 \\
DMASS+DCSS$_{beam=8}$ & 5.43 & 1.19 & 17.83 & 26.29 & 14.08 & 50.62 & 14.17 & \underline{18.89} & \underline{43.54} & \underline{24.47} \\ \hline
\end{tabular}
% \end{adjustwidth}
\caption{Performance of baselines and proposed models on the Newsela dataset.}
\label{tab:perf_newsela}
\end{table*}

In order to make comprehensive comparisons with the state-of-the-art models, we include multiple baselines from the literature, including PBMT-R~\cite{wubben2012sentence}, Hybrid~\cite{narayan2014hybrid}, and SBMT-SARI~\cite{xu2016optimizing}. We also include several strong baselines based on neural networks such as RNN/LSTM, DRESS, DRESS-LS~\cite{zhang2017sentence} as shown in Tables ~\ref{tab:perf_all} and ~\ref{tab:perf_newsela}
% \footnote{More results of evaluation can be seen in the Section 2.2 of Appendix.}.
We developed three models for this experiment. They are DMASS, DCSS, and DMASS+DCSS, where DMASS+DCSS indicates the combination of DMASS and DCSS. The subscript $beam$ indicates the size of beam search.

\paragraph{Results with FKGL Metric}

As shown in Tables ~\ref{tab:perf_all} and ~\ref{tab:perf_newsela}, Hybrid achieves the lowest (thus the best) FKGL score, and DRESS and DRESS-LS have the second best FKGL scores. All the other models including ours do not perform as well as these two. But FKGL measures the simplicity of a sentence without considering the ground truth simplification references, so high FKGL may be at the cost of losing information and readability.

To further analyze the FKGL results, we examine the average sentence length and word length of the outcomes of the models and they are listed as WLen (Word Length) and SLen (Sentence Length) in Tables \ref{tab:perf_all} and ~\ref{tab:perf_newsela}. Hybrid, DRESS, and DRESS-LS are good at generating shorter sentences, but they are not as good at choosing shorter words. In contrast, SBMT-SARI, DCSS, and DMASS all generate shorter words. Therefore, we believe that, by optimizing language model as a goal for the reinforcement learning, DRESS and DRESS-LS are tuned to simplify sentences by shortening the sentence lengths. 
In contrast, with the help of an integrated external knowledge base, SBMT-SARI and our models have more capability to generate shorter words in order to simplify sentences. Therefore, these two sets of models complete sentence simplification tasks via different routes, and perhaps there should be an exploration of combining these two routes for even more successful sentence simplification.

Another interesting finding is that the larger beam search size increases average word length slightly. This is because the larger beam search size mitigates the issue of the inaccurate simplification so that fewer words are simplified.
To measure the correctness of simplification, we analyze the SARI metric and Rule Utilization.

\paragraph{Results with SARI Metric}
\label{sec:sari}

SARI is the most reliable metric for the sentence simplification task~\cite{xu2016optimizing}, therefore we would like to present more detailed discussion regarding the SARI results. As shown in Tables~\ref{tab:perf_all} and ~\ref{tab:perf_newsela}, DMASS+DCSS achieves the best SARI score, which demonstrates the effectiveness of integrating the knowledge base Simple PPDB for sentence simplification.   

To further examine the impacts of the F1 scores for three operations in calculating the SARI scores, as shown in Tables~\ref{tab:perf_all} and ~\ref{tab:perf_newsela}, DMASS+DCSS, as well as other models with high SARI performance benefit greatly by correctly adding and deleting words. We believe these benefits mostly result from the integration with the knowledge base, which provides reliable guidance about which words to modify. 
SBMT-SARI, which represents a previous state-of-the-art model that also integrates with knowledge bases, performs best in correctly adding new words but performs inferiorly in deleting/keeping words. By analyzing the outputs, SBMT-SARI acts aggressively to simplify as many words as possible. But it also results in incorrect simplification. DRESS and DRESS-LS are inclined to generate the shorter sentence, which leads to high F1 scores for deleting words, but it lags behind other models in adding/keeping words.

% it doesn't mean achieving a good performance in adding/keeping words.
% By analyzing the outputs, we found those SBMT-SARI tries to simplify every word in sentence without considering the context. DMASS uses an additional augmented memory to remember the context whereas DCSS encode the context internally. 
% Our models, especially DMASS+DCSS perform well in combination of adding, deleting and keeping words. We believe it benefits from our integrating approach focus more on the context.

DMASS leverages an additional memory component to maintain the simplification rules; DCSS uses internal memory to store those rules. A large number of simplification rules might confuse the model with limited internal memory. This might be the reason why DMASS works better than DCSS.
By taking a two-way advantage of both models, DMASS+DCSS takes a two-fisted approach to store the simplification rules in both additional and internal memory. As a result, DMASS+DCSS achieves the best performance in SARI.

\paragraph{Results with Rule Utilization}
\label{sec:rule}

In this section, we evaluate the models' capabilities for word transformation. The majority of previous approaches, except for the SBMT-SARI, perform poorly in recall. We believe the knowledge base Simple PPDB will reduce uncertainty in the word selection. 

As before, SBMT-SARI acts aggressively to simplify every word in the sentence. Such an aggressive action leads to relatively high performance in recall. However, it does not achieve a strong performance in precision. 
DMASS performs better in terms of rule utilization as compared to DCSS by leveraging an additional memory. 
DMASS+DCSS takes advantage of both approaches that store the simplification rules in additional and internal memory. This combined model is guaranteed to apply more accurate rules.

As compared to the loose relationship between SARI and beam search size, 
we find that that beam search size correlates strongly with the performance in rule utilization. Thus, we believe larger beam search size contributes to good coverage of rules to be applied as well as accuracy in applying rules.

% Through analysis of rule utilization, we found that our model achieves both a good coverage of rules to be applied (recall) and accuracy gained by applying the appropriate rules (precision).
% tries to simplify every word in sentence without considering the context that leads to low performance in precision despite its high performance in recall. Our three models, especially DMASS+DCSS perform relatively good in precision and recall that result in highest performance in F1.

\section{Conclusion}

In this paper, we propose two innovative approaches for sentence simplification based on neural networks. Both approaches are based on multi-layer and multi-head attention architecture and integrated with the Simple PPDB, an external sentence simplification knowledge base, in different ways. By conducting a set of experiments, we demonstrate that the proposed models perform better than existing methods and achieve new state-of-the-art in sentence simplification. 
Our experiments firstly prove that the multi-layer and multi-head attention architecture has an excellent capability to understand the text by accurately selecting specific words in a normal sentence and then choosing right simplified words. Secondly, by integrating with the knowledge base, our models outperform multiple state-of-the-art baselines for sentence simplification.
Compared to previous models which integrated with the knowledge base, our models, especially, DMASS+DCSS, provide both good coverage of rules to be applied and accuracy in applying the correct rules.
In future, we would like to investigate deeper into the different effects of additional memory and internal memory.

\section{Acknowledge}

This research was supported in part by the University of Pittsburgh Center for Research Computing through the resources provided.
The research is funded in part by grants the National Institute on Disability, Independent Living, and Rehabilitation Research (NIDILRR) \#90RE5018 and \#90DP0064, and by Pittsburgh Health Data Alliance's ``CARE'' Project.

\bibliography{emnlp2018}
\bibliographystyle{acl_natbib_nourl}

\appendix

\end{document}